\documentclass{article}

\PassOptionsToPackage{numbers, compress}{natbib}

\usepackage[preprint]{main}

\usepackage[utf8]{inputenc} %
\usepackage[T1]{fontenc}    %
\usepackage{hyperref}       %
\usepackage{url}            %
\usepackage{booktabs}       %
\usepackage{amsfonts}       %
\usepackage{nicefrac}       %
\usepackage{microtype}      %
\usepackage{xcolor}         %
\hypersetup{hidelinks}

\usepackage{multirow}
\usepackage{graphicx}
\usepackage{longtable}
\usepackage{color, colortbl}
\usepackage{appendix}
\usepackage{chngcntr}
\usepackage{geometry}
\usepackage{nicematrix}
\usepackage{algorithm}
\usepackage{ragged2e}
\usepackage{algpseudocode, float}
\usepackage{esvect}
\usepackage{color, colortbl}
\usepackage{float}
\usepackage{afterpage}
\usepackage{amsmath,amssymb}
\usepackage{wrapfig}
\usepackage{caption}
\usepackage{threeparttable}
\usepackage{todonotes}
\usepackage{subcaption}
\usepackage[acronym]{glossaries}
\usepackage{silence}
\usepackage{scalerel}
\makeglossaries

\definecolor{highlight}{RGB}{255,153,0}

\algnewcommand{\Inputs}[1]{%
  \State \textbf{Inputs:}
  \Statex \hspace*{\algorithmicindent}\parbox[t]{.9\linewidth}{\raggedright #1}
}
\algnewcommand{\Initialize}[1]{%
  \State \textbf{Initialize:}
  \Statex \hspace*{\algorithmicindent}\parbox[t]{.9\linewidth}{\raggedright #1}
}
\algnewcommand{\Outputs}[1]{%
  \State \textbf{Outputs:}
  \Statex \hspace*{\algorithmicindent}\parbox[t]{.9\linewidth}{\raggedright #1}
}
\mathchardef\mhyphen="2D

\newcommand{\glslong}[1]{\glsentrylong{#1} (\glsentryshort{#1})}

\newacronym{ann}{ANN}{Artificial Neural Network}
\newacronym{sota}{SOTA}{state-of-the-art}
\newacronym{dnn}{DNN}{Deep Neural Network}
\newacronym{gb}{GB}{Gradient Backpropagation}
\newacronym{hl}{HL}{Hidden Layer}
\newacronym{lrp}{LRP}{Layer-wise Relevance Propagation}
\newacronym{ml}{ML}{Machine Learning}
\newacronym{mlp}{MLP}{Multi Layer Perceptron}
\newacronym{pm}{PM}{Parameter Magnitude}
\newacronym{re}{RE}{Random Erasing}
\newacronym{reldrop}{RelDrop}{Relevance-driven Input Dropout}
\newacronym{sgd}{SGD}{Stochastic Gradient Descent}
\newacronym{xai}{XAI}{eXplainable artificial intelligence}
\newacronym{timm}{timm}{Pytorch Image Models}
\newacronym{gt}{GT}{Ground Truth}
\newacronym{rra}{RRA}{Relevance Rank Accuracy}

\graphicspath{{Figures/}{../Figures/}}

\title{Relevance-driven Input Dropout: an Explanation-guided Regularization Technique}

\author{Shreyas Gururaj$^{1,2}$ \and 
Lars Grüne$^2$ \and
Wojciech Samek$^{1,3,4}$ \and
Sebastian Lapuschkin$^{1,5,\dagger}$ \and
Leander Weber$^{1,\dagger}$ \and
\\
$^1$Fraunhofer Heinrich Hertz Institute, Department of Artificial Intelligence, Berlin, Germany\\
$^2$Lehrstuhl für Angewandte Mathematik, University of Bayreuth, Bayreuth, Germany\\
$^3$ Technische Universität Berlin, Berlin, Germany\\
$^4$ BIFOLD – Berlin Institute for the Foundations of Learning and Data, Berlin, Germany\\
$^5$ Centre of eXplainable Artificial Intelligence, Technological University Dublin, Dublin, Ireland\\
$^\dagger$ Corresponding: \texttt{\{sebastian.lapuschkin,leander.weber\}@hhi.fraunhofer.de}\\
}

\begin{document}
\maketitle

\begin{abstract}

  Overfitting is a well-known issue extending even to \gls{sota} \gls{ml} models, resulting in reduced generalization, and a significant train-test performance gap. Mitigation measures include a combination of dropout, data augmentation, weight decay, and other regularization techniques. Among the various data augmentation strategies, occlusion is a prominent technique that typically focuses on randomly masking regions of the input during training. Most of the existing literature emphasizes \textbf{randomness} in selecting and modifying the input features instead of regions that strongly influence model decisions. We propose \textbf{\glslong{reldrop}}, a novel data augmentation method which selectively occludes the most relevant regions of the input, nudging the model to use other important features in the prediction process, thus improving model generalization through informed regularization. We further conduct qualitative and quantitative analyses to study how \gls{reldrop} affects model decision-making. Through a series of experiments on benchmark datasets, we demonstrate that our approach improves robustness towards occlusion, results in models utilizing more features within the region of interest, and boosts inference time generalization performance. Our code is available at \href{https://github.com/Shreyas-Gururaj/LRP_Relevance_Dropout}{\url{https://github.com/Shreyas-Gururaj/LRP_Relevance_Dropout}}.
  
\end{abstract}

\section{Introduction}

Deep learning models have achieved remarkable success in recent years, demonstrating superior performance in areas such as highly accurate classification \citep{Dosovitskiy2021ViT}, writing of complex text \citep{Ouyang2022InstructGPT}, or assisting in medical diagnosis \citep{Briganti2020Artificial}. Training such models requires vast amounts of (high-quality) data, which may be unavailable in real-world applications due to high annotation costs. Consequently, overfitting is a common occurrence, where biases or regularities specific to the training set are reflected by the model, resulting in bad generalization to unseen examples. However, as acquiring more training data tends to be difficult and expensive, eliminating such regularities by including additional (counter-)examples for training is generally infeasible.

For this reason, data augmentation techniques are commonly employed to mitigate overfitting. These techniques introduce variations in training data, thereby artificially enlarging the available data and effectively regularizing the model to learn more robust representations. However, traditional data augmentation techniques are either computationally demanding \cite{Cubuk2019AutoAugment}, subject to the same regularities as the training data \cite{Bowles2018GANAugmentation}, or based on random perturbations \cite{SinghL17,KangDZY17,ZhangCDL18,Inui2019EDA,YunHCOYC19,Zhong0KL020}, which may not align with the parts of the data that are problematic in terms of overfitting. At worst, models may overfit on the same features as on unaugmented data if these features are perturbed as frequently as non-problematic features. 

Here, techniques from the domain of \gls{xai} offer a solution by providing importance scores that offer insights into which features contribute to model predictions. Previous works have successfully applied these importance scores to improve otherwise random regularization techniques within the model \cite{Zunino2021Excitation,YangDQWN22}. Consequently, leveraging the information provided by attributions in the context of data augmentation by considering the importance of different input regions in a more principled manner seems promising.

We therefore introduce \glslong{reldrop}, a novel \gls{xai}-guided technique that augments data by masking (currently) relevant input features. Our method can be efficiently applied during training, requiring only one additional backward pass per batch. We demonstrate its effectiveness in the context of image and point cloud classification and validate the increased robustness and generalization ability of the resulting model. In summary, we make the following contributions:
\begin{itemize}
    \item We introduce \gls{reldrop}, a novel data augmentation technique that utilizes information from attributions to mask input features in a principled manner.
    \item We discuss the advantages, limitations, and caveats of utilizing \gls{xai} to guide input perturbations. 
    \item We validate empirically that \gls{reldrop} increases the model's ability to generalize to unseen data in both image and point cloud classification domains, compared to the baselines of no input data augmentation or random input data augmentation.
    \item We investigate the effects of \gls{reldrop} on model robustness qualitatively and quantitatively, demonstrating that it leads to the models utilizing more features for inference and growing more resilient against feature removal, effectively counteracting overfitting.
\end{itemize} 

\begin{figure}
\centering
\includegraphics[width=\textwidth]{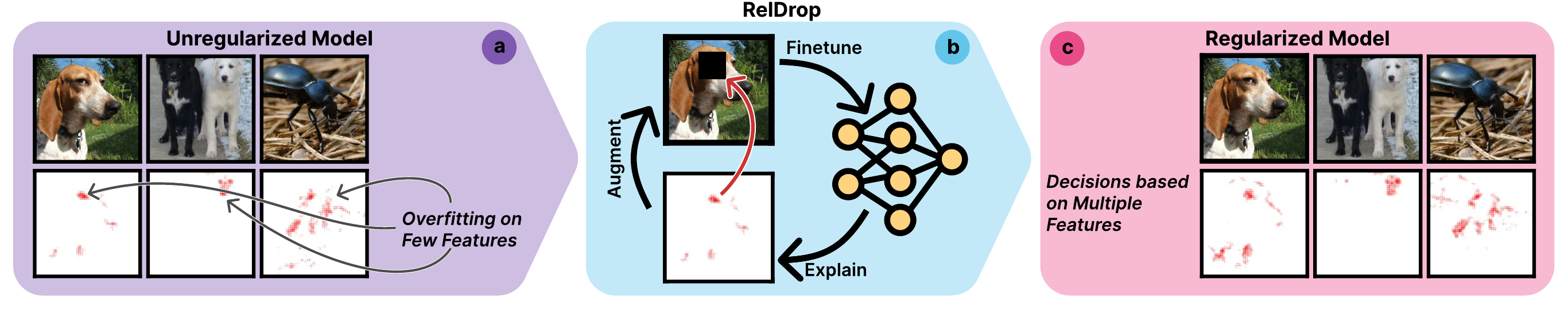}
\caption{Schematic overview of \gls{reldrop} workflow, illustrated for image data. As shown by the relevance heatmaps, Panel (a) depicts an \emph{Unregularized Model} that, without data augmentation, overfits by heavily depending on a few highly relevant features. To nudge the model into learning from a larger set of informative features, \emph{\gls{reldrop}} repeatedly computes attributions during training, and then augments the data by selectively masking the (currently) most relevant parts of the input (Panel (b)). The impact of training with \gls{reldrop} is clear in Panel (c), where heatmaps at the bottom reflect how the \emph{Regularized Model} utilizes a larger set of features to make predictions.}
\label{figure:reldrop_overview}
\end{figure}

\section{Related Work}
\label{section:RelatedWork}
\label{subsection:Regularization}
Regularization in the context of a \gls{dnn} refers to the various techniques that aim to improve model generalization ability by reducing overfitting. The most common regularization techniques can be broadly classified into dropout \citep{abs-1207-0580,SrivastavaHKSS14,LiangWLWMQCZL21,WanZZLF13,BaF13,abs-1301-3557,XieWWWT16}, data augmentation \citep{KangDZY17,Zhong0KL020,YunHCOYC19,SinghL17,ZhangCDL18,YangDQWN22}, weight decay, \citep{KroghH91,10.1111/j.1467-9868.2005.00503.x,LoshchilovH19} and noise injection \citep{CamutoWSRH20,MullerKH19,NohYMH17}.

For any mapping function to be considered generalizing, it not only map the training samples precisely, but also any additional samples from the (source) data distribution \citep{KondaBMV15}. Geometric transformations, such as random flipping, rotation, and cropping, are often applied during training \citep{Krizhevsky2012Imagenet, He2016Deep}. More recent and advanced techniques such as \gls{re} \citep{Zhong0KL020}, Mixup \citep{ZhangCDL18}, CutMix \citep{YunHCOYC19}, Hide-and-Seek \citep{SinghL17}, and PatchShuffle \citep{KangDZY17} are a class of \emph{random} occlusion-based data augmentations applied to inputs along with their labels.
However, models regularized by the above methods may overfit on the same features, if the randomly occluded features are not (or rarely, over the training process) the ones being overfit on. For this reason, we propose to leverage \gls{xai} attributions as a signal to guide data augmentation, as they provide a more informed approach to examine how these augmentations affect model predictions.

Originally, the field of \gls{xai} research aims to reveal the prediction mechanisms underlying otherwise black-box models. \emph{Post-hoc} \gls{xai} techniques seek to explain already trained models and can be broadly classified into two types: \emph{global} explanations, which are overall insights into model behavior (e.g., discovery of encoded concepts or learned representations \citep{Bau2017Network,Kim2018Interpretability,Hu2020Architecture,Hohman2020Summit,Achtibat2023CRP,Vielhaben2023Multi,Huben2024SAE,Gorton2024Missing,Bhalla2024Interpreting,Kowal2024Understanding,Dreyer2025SemanticLens}), and \emph{local} explanations, which provide insights into individual predictions by assigning importance to input features, e.g., via attribution scores. Sampling-based techniques \citep{Strumbelj2014Explaining,Ribeiro2016Why, Zintgraf2017Visualizing,Fong2017Interpretable,Lundberg2017Unified} are computationally intensive but treat the model as a black box. In contrast, backpropagation-based methods \citep{Baehrens2010How,Bach2015Pixel,Montavon2017Explaining,Shrikumar2017Learning,Sundararajan2017Axiomatic} require internal access to model parameters but are efficient, needing only a single forward and backward pass. Alongside providing human-understandable input-level explanations, several backpropagation-based \gls{xai} methods provide intermediate importance scores for internal activations or weights. Consequently, they have been employed for applications such as pruning, quantization, and freezing of network components \citep{Voita2019Analyzing,Sabih2020Utilizing,Yeom2021Pruning,BeckingDSML20,Hatefi2024Pruning,Ede2022Explain}, or as an auxiliary training signal for gradient masking \citep{Lee2019Improvement,Nagisetty2020xai} and credit assignment \citep{Weber2025Efficient}. Among these, some approaches \citep{Zunino2021Excitation,YangDQWN22} leverage \gls{xai} to control information flow in intermediate layers via explanation-based filtering. \gls{reldrop} is related to these (due to certain equivalencies between \emph{dropout} and \emph{data augmentation}, cf. \citep{ZhaoYXL19,KondaBMV15,abs-2103-15027}), but is applied directly to the input, targeting the most salient regions and selectively masking them. Consequently, any information occluded by \gls{reldrop} is completely removed from the forward pass (as opposed to occlusions in feature space, that allow for the occluded information to remain available to the network, via a different path).

\section{RelDrop} 
\label{section:RelDrop}
This Section introduces \gls{reldrop} and discusses how it is used for 2D image classification and 3D point cloud classification tasks. Our approach draws inspiration from data augmentation techniques based on random occlusion methods for 2D images \citep{Zhong0KL020, ZhangCDL18, YunHCOYC19, SinghL17, KangDZY17} and 3D point cloud domains \citep{QiSMG17, QiYSG17, WangSLSBS19, GuoCLMMH21}, respectively. These methods generally augment random input regions without taking into account how the augmented features affect model predictions. Therefore, the likelihood of augmenting a particular feature that the model has overfitted on is as likely as augmenting a particular feature that the model is invariant to. In comparison, \gls{reldrop}, as shown schematically in Figure \ref{figure:reldrop_overview}, masks input regions based on the respective attribution maps and serves as an input regularizer that mitigates overfitting by occluding features that are currently relevant to the model prediction. For this purpose, we utilize an attribution function $\mathcal{R}$ that assigns importance scores to all features of the input $\mathcal{I}$. The most important features, as indicated by the attribution map, are consequently masked. We define the \gls{reldrop} input augmentation as follows:

\begin{equation} 
    \label{eqn:RelDrop}
    \mathcal{I}_\text{{\gls{reldrop}}} = \mathcal{M}_\mathcal{R} \odot \mathcal{I} + (1 - \mathcal{M}_\mathcal{R}) \cdot \mathbf{s}
\end{equation} 

Here, $\odot$ denotes element-wise multiplication. $\mathcal{I}$ is the original input, and $\mathcal{M_\mathcal{R}}$ is a binary mask that occludes input regions based on attributions $\mathcal{R}$, which assigns importance scores to input features. In practice, we use \gls{lrp} \cite{Bach2015Pixel} for attribution due to its  pixel- or point-level granularity, computational efficiency, and prior success in several similar applications \cite{Yeom2021Pruning,Ede2022Explain,Weber2025Efficient}. The replacement value for all the masked features $\mathbf{s}$ (e.g., dataset mean or zero) can vary with the data modality.
Equation \ref{eqn:RelDrop} is a general formulation of \gls{reldrop} that can be adapted to different data modalities, such as 2D images and 3D point clouds.

To balance performance and regularization, controlling how and when input features are masked is important. We hypothesize that persistent occlusion of the features that a model has learned to be most informative (i.e., that have the largest attribution scores) may be detrimental to convergence. This hypothesis aligns with previous work \citep{YangDQWN22}, where the authors note that aggressive removal of high-attribution positions negatively impacts convergence. We therefore implement a balanced masking strategy that only drops high-attribution features with a large probability, as this strongly encourages the model to utilize other features while preserving core information needed for stable optimization. Furthermore, unlike dropout \citep{SrivastavaHKSS14} of internal activations, where the (complete) input is still encoded by preceding layers and alternate paths, masking input features removes information entirely, making it unavailable to the model. Excessive masking in input space (e.g., with values common for dropout, such as $50\%$) may therefore cause the model to train on an insufficient signal, degrading performance. 

For the above reasons, we design the binary mask function $\mathcal{M_\mathcal{R}}$ to retain partial randomness and control the proportion of input feature replacement via tunable hyperparameters, ensuring not all relevant information is masked. For 2D images, we utilize the same hyperparameters as \gls{re} \cite{Zhong0KL020}: dropout probability $p$ to control the augmentation frequency and occlusion area $S_O$ combined with aspect ratio $r_O$ to affect the occluded area. For 3D point clouds, we introduce two similar hyperparameters: $\alpha$ for the proportion of attribution-guided vs. randomly applied augmentation and $\beta$ for the overall fraction of points replaced. The exact use of these hyperparameters is detailed in the following sections.

\subsection{2D Image Classification} \label{subsection:RelDrop2D}
For 2D image classification, the input is an image $\mathcal{I}^{2d} \in \mathbb{R}^{H \times W \times C}$, where $H$ and $W$ are the height and width of the image, respectively, and $C$ the number of channels (e.g., $C = 3$ for RGB images). In this setting, we extend \gls{re} \citep{Zhong0KL020} by centering the occlusion region around the most important pixel, as determined by the normalized attribution $\mathcal{R}_{\text{norm}}^{2d} \in \mathbb{R}^{H \times W}$, instead of randomly selecting it. Let $S=H \times W$ be the area of the image, and $(x, y)$ denote the height and width coordinates of a pixel of $\mathcal{I}^{2d}$. The normalized relevance map \( \mathcal{R}_{norm}^{2d} \) is obtained by summing channel-wise over the original map \( \mathcal{R} \in \mathbb{R}^{C \times H \times W} \)  and then normalizing the result to be interpretable as input dropout probabilities, bounded in [0, 1]:
\begin{gather}
    \mathcal{R}^{2d}(H, W) = \sum_{c=1}^{C} \mathcal{R}(c, H, W), \\
    \mathcal{R}_{\text{norm}}^{2d} = 
    \frac{\mathcal{R}^{2d}(H, W) + \max\left(|\mathcal{R}^{2d}(H,W)|\right)}
         {2 \cdot \max\left(|\mathcal{R}^{2d}(H,W)|\right)} \in [0, 1]
    \label{eqn:RelDrop2DNorm}
\end{gather}

We then select the centroid pixel to be the most relevant pixel:

\begin{equation} 
    \label{eqn:RelDrop2DCentroid}
    (x_{\text{cen}}, y_{\text{cen}}) = \arg\max_{(x, y)} \mathcal{R}_\text{norm}^{2d}
\end{equation}

Similar to \citet{Zhong0KL020}, we occlude a rectangular block around $(x_{\text{cen}}, y_{\text{cen}})$, with dimensions $H_O$ and $W_O$ computed as:

\begin{equation}
    \begin{gathered}
        H_O = \sqrt{S_O \cdot r_O}, \quad W_O = \sqrt{\frac{S_O}{r_O}},
        \\
        \text{where} \quad S_O = U(S_{\text{low}}, S_{\text{high}}) \cdot S, \quad r_O = U(r_{\text{low}}, \frac{1}{r_{\text{low}}})
    \end{gathered}
\end{equation}

Here, $U(\cdot)$ represents a random uniform distribution defined by the hyperparameters $S_{\text{low}}$, $S_{\text{high}}$ and $r_{\text{low}}$. $S_O$ denotes the area of the occlusion region (i.e., the number of pixels to be masked), and $r_O$ is the aspect ratio, determining the relative dimensions of the height and width of the rectangular block. The occlusion region $O$ is then determined by width $W_O$ and height $H_O$:

\begin{equation} \label{eqn:RelDrop2DRegion}
O = \left\{ (x, y) \ | \ 
x \in \left[ x_{\text{cen}} - \frac{W_O}{2}, x_{\text{cen}} + \frac{W_O}{2} \right],
y \in \left[ y_{\text{cen}} - \frac{H_O}{2}, y_{\text{cen}} + \frac{H_O}{2} \right]
\right\}
\end{equation}

Since only pixels in $O$ are occluded, the occlusion mask is defined as follows:

\begin{equation} \label{eqn:RelDrop2Dmask}
\mathcal{M}^{2d}_\mathcal{R} = \begin{cases} 0, & \text{if} \ (x, y) \in O \\ 1, & \text{otherwise} \end{cases} 
\end{equation} 

Finally, we apply $\mathcal{M}_\mathcal{R}^{2d}$ to the input image, according to Equation \eqref{eqn:RelDrop}, choosing the channel-wise dataset mean $\mu$ as the replacement value $\mathbf{s}$. A pseudo-algorithm inspired by \gls{re} \citep{Zhong0KL020}, which depicts the application of \gls{reldrop} to an input image is shown in the Appendix \ref{appendix:AlgRelDrop2D}

\subsection{3D Point Cloud Classification}
\label{subsection:RelDrop3D}
In this setting, an input consists of a point cloud $\mathcal{I}^{3d} = \{(x_i, y_i, z_i) | i \in \{1, ..., N\}\}$, i.e., a collection of $N$ points $(x_i, y_i, z_i) \in \mathbb{R}^3$, encoding spatial coordinates. %
Similarly, an importance score is attributed per point as a vector $\mathcal{R}^{3d} \in \mathbb{R}^N$. In contrast to 2D image data, where we adopted block-level operations for input occlusion, we instead replace individual points from $\mathcal{I}^{3d}$ with the origin, setting them to $(0, 0, 0)$ (cf. Equation \eqref{eqn:RelDrop}: $s=(0, 0, 0)$) in the 3D setting. This nullifies their influence on the prediction, as the PointNet++ model \citep{QiYSG17} is designed to operate on a metric loss and to generate a fixed-size embedding for a given input point cloud of fixed size $N$. Additionally, we explicitly balance random and attribution-based modifications to prevent the complete removal of all the important features necessary for making a prediction. For this purpose, we introduce parameters $\alpha, \beta \in  [0, 1]$, where 

\begin{itemize}
    \item $\alpha$ determines the proportion of random to attribution-guided input data augmentation.
    \item $\beta$ controls the total percentage of occluded points.
\end{itemize}

To compute the mask $\mathcal{M}_\mathcal{R}^{3d}$, we again normalize the intermediate relevance to be interpretable as input dropout probabilities:
\begin{equation} 
    \label{eqn:normalize}
    \mathcal{R}^{3d}_\text{norm} = \frac{\mathcal{R}^{3d} - \min(\mathcal{R}^{3d})}{\max(\mathcal{R}^{3d}) - \min(\mathcal{R}^{3d})} \quad \mathcal{R}_{\text{norm}}^{3d} \in [0, 1]
\end{equation}

Using this $\mathcal{R}^{3d}_\text{norm}$, the mask is then generated as follows:

\begin{equation} 
    \label{eqn:RelDrop3Dmask}
    \mathcal{M}_\mathcal{R}^{3d}  = \begin{cases}
        0, & \text{if } \alpha \vv{v} + (1 - \alpha) \mathcal{R}^{3d}_\text{norm} \geq (1 - \beta) \\
        1, & \text{otherwise}
    \end{cases}
\end{equation}

where ${\vv{v}} \in [0.0, 1.0]^N$ is a 1D vector of size $N $ that represents random masking, sampled from a uniform distribution. Finally, we apply the mask to the input point cloud by setting the points corresponding to \( \mathcal{M}_\mathcal{R}^{3d}  = 0 \) to the origin.

\section{Results}
\label{section:results}
In this section, we study the effects of applying \gls{reldrop} for 2D image classification and 3D point cloud classification tasks, respectively. Refer to Appendices \ref{appendix:datasets}, \ref{appendix:metrics}, and \ref{appendix:exp} for details on the chosen benchmark datasets, evaluation metrics, experimental setup, and hyperparameters, respectively.

\definecolor{group1}{RGB}{173, 216, 230} %
\definecolor{group2}{RGB}{144, 238, 144} %
\definecolor{group3}{RGB}{255, 223, 186}

\begin{table}[h]
    \centering
    \small
     \caption{Effects of \gls{reldrop} on ResNet generalization ability. We evaluate the test accuracies (\%, on the test and validation sets, respectively)  of models trained on CIFAR-10/100 and ImageNet-1k datasets (\textbf{\textcolor{group1}{Finetuning}}) and the zero-shot test accuracy (\%) of ImageNet-1k trained models on ImageNet-R \citep{HendrycksBMKWDD21}, ImageNet-A, and ImageNet-O \citep{HendrycksZBSS21} datasets (\textbf{\textcolor{group2}{Zero-shot}}). During training, we employ different input augmentation schemes, where \emph{RE} and \emph{\gls{reldrop} (ours)} represent \gls{re} \citep{Zhong0KL020} and our method, respectively, and \emph{baseline} denotes an augmentation-free setting. For CIFAR-10/100, the mean and standard deviation over 3 different random seeds are reported and for ImageNet-1k, results are reported for a single randomly chosen seed. The values highlighted in bold represent the best-performing model for the respective model and dataset. Also, the Mean Relevance Rank Accuracy (RRA) (\%) \citep{ArrasOS22} computed over 12419 test samples from in the ImageNet-S validation set, along with their respective segmentation maps \citep{GaoLYCHT23}, is shown in the rightmost column (\textbf{\textcolor{group3}{Mean RRA (\%)}}). \gls{reldrop} improves upon the estimated model generalization ability in all investigated settings compared to \gls{re} and the baseline and increases \emph{Mean RRA}, except for \emph{Mean RRA}.}

    \resizebox{\textwidth}{!}{%
    \begin{tabular}{c c | c c c | c c c | c }
    \hline
    \multirow{2}{*}{Model} & Augmentation & \multicolumn{3}{c|}{\cellcolor{group1} Finetuning (test Acc)} & \multicolumn{3}{c|}{\cellcolor{group2} Zero-shot (test Acc)} & \multicolumn{1}{c}{\cellcolor{group3} Mean RRA} \\
    \cline{3-9}
    & Type & CIFAR-10 & CIFAR-100 & ImageNet-1k & ImageNet-R & ImageNet-A & ImageNet-O & ImageNet-S \\
    \hline
    \multirow{3}{*}{ResNet-18} 
    & baseline & 94.98$_{\pm 0.08}$ & 76.80$_{\pm 0.19}$ & 71.39 & 32.10 & 1.67 & 15.66 & 59.68 \\
    & RE \citep{Zhong0KL020} & 95.27$_{\pm 0.07}$ & 76.90$_{\pm 0.24}$ & 71.54 & 32.34 & 1.72 & 15.57 & 59.72 \\
    & RelDrop (Ours) & \textbf{95.70}$_{\pm 0.04}$ & \textbf{77.41}$_{\pm 0.14}$ & \textbf{71.91} & \textbf{33.15} & \textbf{1.91} & \textbf{15.67} & \textbf{60.02} \\
    \hline
    \multirow{3}{*}{ResNet-34} 
    & baseline & 95.63$_{\pm 0.13}$ & 78.84$_{\pm 0.31}$ & 76.84 & 36.18 & 3.95 & 16.94 & 58.83 \\
    & RE \citep{Zhong0KL020} & 96.10$_{\pm 0.07}$ & 79.33$_{\pm 0.27}$ & 77.04 & 36.07 & 3.97 & 16.95 & 59.19 \\
    & RelDrop (Ours) & \textbf{96.30}$_{\pm 0.09}$ & \textbf{79.56}$_{\pm 0.19}$ & \textbf{77.64} & \textbf{37.17} & \textbf{4.05} & \textbf{17.01} & \textbf{59.54} \\
    \hline
    \multirow{3}{*}{ResNet-50} 
    & baseline & 95.41$_{\pm 0.12}$ & 79.01$_{\pm 0.35}$ & 79.19 & 37.85 & 8.37 & 19.32 & 61.74 \\
    & RE \citep{Zhong0KL020} & 95.63$_{\pm 0.14}$ & 79.72$_{\pm 0.42}$ & 79.45 & 37.88 & 8.69 & 19.22 & 62.23 \\
    & RelDrop (Ours) & \textbf{95.93}$_{\pm 0.11}$ & \textbf{80.35}$_{\pm 0.31}$ & \textbf{79.94} & \textbf{38.65} & \textbf{9.17} & \textbf{19.37} & \textbf{62.47} \\
    \hline
    \multirow{3}{*}{ResNet-101} 
    & baseline & - & - & 80.84 & 42.31 & 16.61 & 22.39 & 56.89 \\
    & RE \citep{Zhong0KL020} & - & - & 81.18 & 42.14 & 17.02 & 22.45 & \textbf{57.91} \\
    & RelDrop (Ours) & - & - & \textbf{81.74} & \textbf{42.75} & \textbf{18.08} & \textbf{22.66} & 57.72 \\
    \end{tabular}
    }
    \label{table:ResNet_results_table_Image}
\end{table}

\subsection{2D Image Classification}
\label{section:2D_Classification}
We first investigate the effect of \gls{reldrop} on model performance in the setting of image classification by comparing the test accuracies of \gls{reldrop}-trained models to that of \gls{re}-trained models and a baseline without data augmentation. We apply block-level occlusions similar to \gls{re}, (cf. the \gls{reldrop} formulation from Section \ref{subsection:RelDrop2D}), and use the pre-trained weights downloaded from the \gls{timm} library \citep{rw2019timm}) as a starting point, which is further finetuned with the different augmentation strategies. We finetune the models for $100$ epochs for CIFAR-10/100 and $50$ epochs for ImageNet-1k, respectively. For \gls{reldrop} and \gls{re}, we set the dropout probability to $p=0.5$ as recommended by the authors of \gls{re} \citep{Zhong0KL020}.

\begin{figure}[h]
    \centering
    \includegraphics[width=\textwidth, height=1.0\textheight, keepaspectratio]{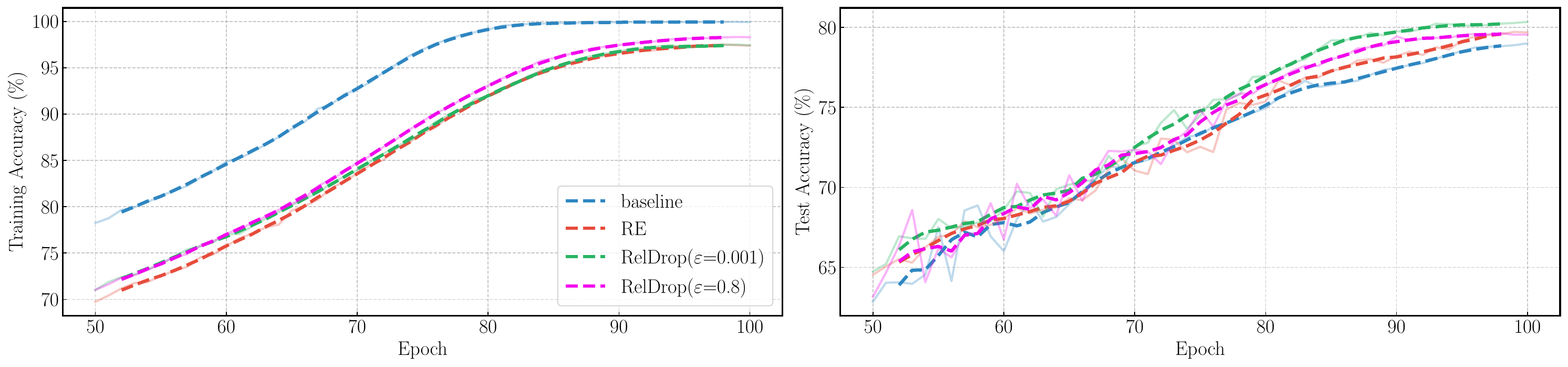}  \caption{Regularization effect of \gls{reldrop} on ResNet50 overfitting. The curves show the augmentation-free baseline (\emph{baseline}), \gls{re} (\emph{\gls{re}}), and two variations of \gls{reldrop} (\emph{\gls{reldrop}}), with attribution hyperparameters $\varepsilon=0.8$ (performing worst) and $\varepsilon=0.001$ (performing best) on the CIFAR-100 dataset. A moving average (more solid lines) with window size $5$ is visualized along with the raw data for both the training (\emph{left}) and test (\emph{right}) curves. \gls{reldrop}-trained model generalizes better to the test data compared to both \gls{re} and the baseline. Note that the y-axis scale of training (\emph{left}) and test (\emph{right}) figures are adjusted to the respective ranges of training and test accuracies.}
    \label{fig:Cifar100_ResNet50}
\end{figure}

As shown in Table \ref{table:ResNet_results_table_Image}  \emph{(Blue Columns)}, models trained with \gls{reldrop} consistently outperform both the \gls{re} and the baseline. Our proposed approach further improves the test performance over the \gls{re} by almost doubling the gain over the baseline in all the considered models and datasets. The average margin of improvement across all the considered models over the baseline is $0.64\%$, $0.89\%$, and $0.93\%$ for CIFAR-10, CIFAR-100, and ImageNet-1k datasets, respectively. This indicates that the regularization effect of \gls{reldrop} results in consistent improvement in the model's generalization ability compared to both \gls{re} and the baseline.

Investigating this effect in more detail in Figure \ref{fig:Cifar100_ResNet50}, we observe that \gls{reldrop}-trained models have the smallest difference between training and test accuracies, indicating reduced overfitting, while still converging similarly as \gls{re}. While the stabilizer hyperparameter $\varepsilon$ introduced by \gls{lrp} affects these results (we investigated $\varepsilon=[0.1,0.01,0.001,0.2,0.4,0.8,1.0,2.0]$, best and worst are shown in the figure), the worst case performance remains close to \gls{re}.

\subsubsection{Generalization in Zero-Shot Applications}
\label{subsection:Mean_RRA}

While these consistent test performance improvements (cf. Table \ref{table:ResNet_results_table_Image} \emph{(Blue Columns)}) suggest that the \gls{reldrop}-trained models generalize better to the test (or validation) data, a better metric to gauge the true generalization would be to validate whether these improvements are further transferred while conducting zero-shot evaluations on datasets with different distributions. Towards this, the best states of all the models finetuned on ImageNet-1k for 50 epochs, while applying both \gls{re} and \gls{reldrop} are used as a starting point for Zero-shot classification of ImageNet-R \citep{HendrycksBMKWDD21}, ImageNet-A, and ImageNet-O \citep{HendrycksZBSS21} test samples and their respective results are reported in the Table \ref{table:ResNet_results_table_Image} \emph{(Green Columns)}. 

These variations of the ImageNet-1k dataset are specifically designed to introduce adversarial samples, distribution shifts, harder-to-recognize objects, and scenarios where models rely on spurious correlations. Evaluating Zero-shot test accuracy on these datasets provides a comprehensive assessment of the model’s robustness, generalization ability, and resistance to distribution shifts. It also signifies the model’s ability to capture semantically meaningful features, mitigate reliance on spurious correlations, and ensure well-calibrated confidence scores when encountering out-of-distribution samples. The performance improvement on ImageNet-1K test data is also observed on the test sets of these ImageNet-1k variants when predicting classes in a zero-shot fashion. 

\begin{figure}[h]
    \centering
    \includegraphics[width=0.85\linewidth]{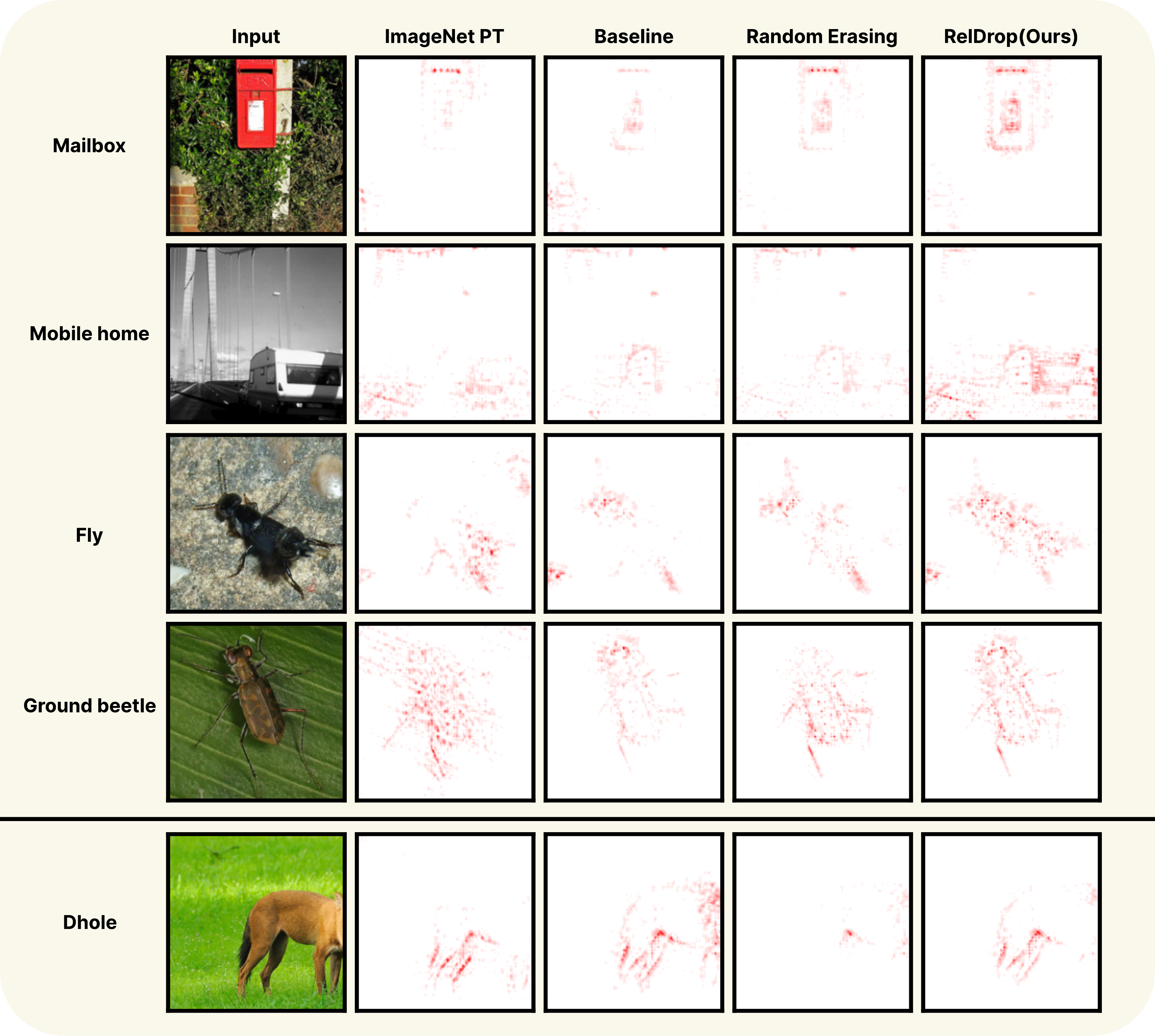}  \caption[Qualitative comparison of baseline, \gls{re} and \gls{reldrop}]{Qualitative effects of \gls{reldrop} on model decision-making. \gls{lrp} attributions are visualized (deeper red indicates higher importance). Columns show, from left to right, the ImageNet pre-trained ResNet50 model, the baseline finetuned without input augmentation, and the models finetuned with \gls{re} and \gls{reldrop}, respectively. The class labels for the input images are shown on the left. We observe increased reliance on within-object features for the \gls{reldrop} finetuned model. Although this effect does not hold for all samples equally (last row), we confirm an overall improvement via the quantitative evaluation of \gls{rra} in Table \ref{table:ResNet_results_table_Image}.}
    \label{fig:Image_Qual_results}
\end{figure}

As shown in Table \ref{table:ResNet_results_table_Image} \emph{(Green Columns)} and the qualitative results in Figure \ref{fig:Image_Qual_results}, \gls{reldrop} encourages the model to learn more robust, albeit not fundamentally different, features across the entire training set, leading to more effective knowledge transfer. \gls{reldrop} produces an average relative improvement over baseline considered across all the models is $2.88\%$, $8.83\%$, and $0.63\%$ for ImageNet-R, ImageNet-A, and ImageNet-O test datasets, respectively (Also, c.f. Appendix \ref{appendix:RelAUCResults}, which indicates that \gls{reldrop} enables the models to learn semantically richer representations). Although these improvements may seem small, they are consistent, with \gls{reldrop} improving the model's zero-shot capabilities over all the considered datasets as opposed to \gls{re}.

In the previous paragraphs, we investigated the effects of \gls{reldrop} on (estimated) model generalization ability. However, \gls{reldrop} functions by removing the input features that are (currently) most relevant to a model's decision-making. As such, while we expect model decisions to be based on a larger number of features after applying \gls{reldrop}, there is no guarantee that these features are part of the object of interest itself. In the worst case, this could lead to models that rely on spurious correlations or context rather than the object itself.

We therefore assess whether \gls{reldrop}-trained models utilize the object of interest for their decision-making by evaluating attribution \gls{rra} \citep{ArrasOS22}. This metric measures the percentage of top positive attributions within a given \gls{gt} mask for each input. Using ImageNet-1k finetuned models with different masking strategies, we compute the \emph{Mean \gls{rra}} over 12,419 test samples found in the validation set of ImageNet-S along with their respective segmentation maps \citep{GaoLYCHT23}, which we employ as \gls{gt} masks. We evaluate the baseline-finetuned models and the best-performing models with \gls{re} and our method, respectively. The results are included in Table \ref{table:ResNet_results_table_Image} \emph{(Orange Columns)}, and show that \gls{reldrop} preserves a similar \emph{mean \gls{rra}} to the baseline and \gls{re}, indicating that no detrimental change in decision-making occurs. Interestingly, we observe that \gls{reldrop} even improves \gls{rra} across all ResNet models except ResNet101, implying less reliance on background features. We hypothesize that by occluding the most relevant features, \gls{reldrop} mitigates the influence of confounders, especially early on during training, while forcing the model to learn more features of the object of interest for several epochs and multiple augmentations.

Additionally, we evaluate the differences in decision-making qualitatively in Figure \ref{fig:Image_Qual_results}. Here, we visualize heatmaps comparing relevance distributions of the original ImageNet-1k pre-trained model, the finetuned baseline (finetuned without data augmentation), and models finetuned while applying \gls{re} and \gls{reldrop}. The first four rows show improved feature distribution for classes \emph{"Mailbox"}, \emph{"Mobile house"}, \emph{"Fly"}, and \emph{"Ground beetle"} while the fifth row highlights an example of \emph{"Dhole"} where our method fails to improve the feature distribution.
While the improved decision-making does not seem to hold for every example, the mean \gls{rra} quantifies this effect on a dataset level, demonstrating the overall effectiveness of our approach.

\subsection{3D Point Cloud Classification}

\begin{table}[t]
    \centering
    \caption[PointNet++ performance on benchmark datasets ModelNet40 and ShapeNet with different dropout parameters]{shows the effect of RelDrop on PointNet++ \citep{QiYSG17} generalization using test accuracy on ModelNet40 \citep{WuSKYZTX15} and ShapeNet \citep{ChangFGHHLSSSSX15}. Models with input augmentation, especially with a balanced strategy replacement ($\alpha$=0.5), outperform the baseline and fully random ($\alpha$=1.0). Reported values are averaged over 3 different random seeds, with their maximum standard deviation, and the bold indicates the best results.} 
    \begin{tabular}{lcccc}
        \hline
        \multirow{2}{*}{\textbf{Dropout}} & \multicolumn{2}{c}{\textbf{ModelNet40}} & \multicolumn{2}{c}{\textbf{ShapeNet}} \\
        \cline{2-5}
        & Inst Acc & Class Acc & Inst Acc & Class Acc \\
        \hline
        baseline & 91.94$_{\pm 0.39}$ & 89.18$_{\pm 0.59}$ & 98.22$_{\pm 0.32}$ & 98.01$_{\pm 0.43}$ \\
        $\alpha$=0.0, $\beta$=0.15 & 92.34$_{\pm 0.49}$ & \textbf{89.92}$_{\pm 0.45}$ & \textbf{98.85}$_{\pm 0.28}$ & 98.30$_{\pm 0.54}$ \\
        $\alpha$=0.0, $\beta$=0.5 & 91.75$_{\pm 0.23}$ & 87.95$_{\pm 0.37}$ & 98.52$_{\pm 0.58}$ & 97.05$_{\pm 0.41}$ \\
        $\alpha$=0.0, $\beta$=0.85 & 64.79$_{\pm 0.37}$ & 55.74$_{\pm 0.46}$ & 95.89$_{\pm 0.37}$ & 85.43$_{\pm 0.29}$ \\
        $\alpha$=0.5, $\beta$=0.15 & \textbf{92.41}$_{\pm 0.38}$ & 89.71$_{\pm 0.59}$ & 98.79$_{\pm 0.27}$ & \textbf{98.62}$_{\pm 0.18}$ \\
        $\alpha$=0.5, $\beta$=0.5 & 92.30$_{\pm 0.12}$ & 89.65$_{\pm 0.24}$ & 98.66$_{\pm 0.31}$ & 97.24$_{\pm 0.55}$ \\
        $\alpha$=0.5, $\beta$=0.85 & 59.73$_{\pm 0.57}$ & 53.24$_{\pm 0.42}$ & 95.67$_{\pm 0.32}$ & 88.99$_{\pm 0.45}$ \\
        $\alpha$=1.0, $\beta$=0.15 & 91.74$_{\pm 0.56}$ & 89.07$_{\pm 0.46}$ & 98.73$_{\pm 0.30}$ & 97.97$_{\pm 0.45}$ \\
        $\alpha$=1.0, $\beta$=0.5 & 91.52$_{\pm 0.55}$ & 89.34$_{\pm 0.34}$ & 98.46$_{\pm 0.44}$ & 98.05$_{\pm 0.39}$ \\
        $\alpha$=1.0, $\beta$=0.85 & 91.13$_{\pm 0.49}$ & 88.65$_{\pm 0.46}$ & \textbf{98.26}$_{\pm 0.61}$ & \textbf{97.59}$_{\pm 0.76}$ \\
        \hline
    \end{tabular}
    \label{table:PointNet++_results_table}
\end{table}

After demonstrating the efficacy of \gls{reldrop} for improving model generalization ability for 2D image classification, we investigate our method's applicability to 3D point cloud classification in the following. For this purpose, we consider the PointNet++ architecture \citep{QiYSG17}, as well as the ModelNet40 and ShapeNet \citep{WuSKYZTX15,ChangFGHHLSSSSX15} benchmark datasets. Unlike the 2D image classification, where block-level occlusions are applied, we augment individual points here (cf. the \gls{reldrop} formulation from Section \ref{subsection:RelDrop3D}) and train models from scratch for 50 epochs instead of fine-tuning.

\begin{figure}[t]
    \centering
    \includegraphics[width=0.7\linewidth]{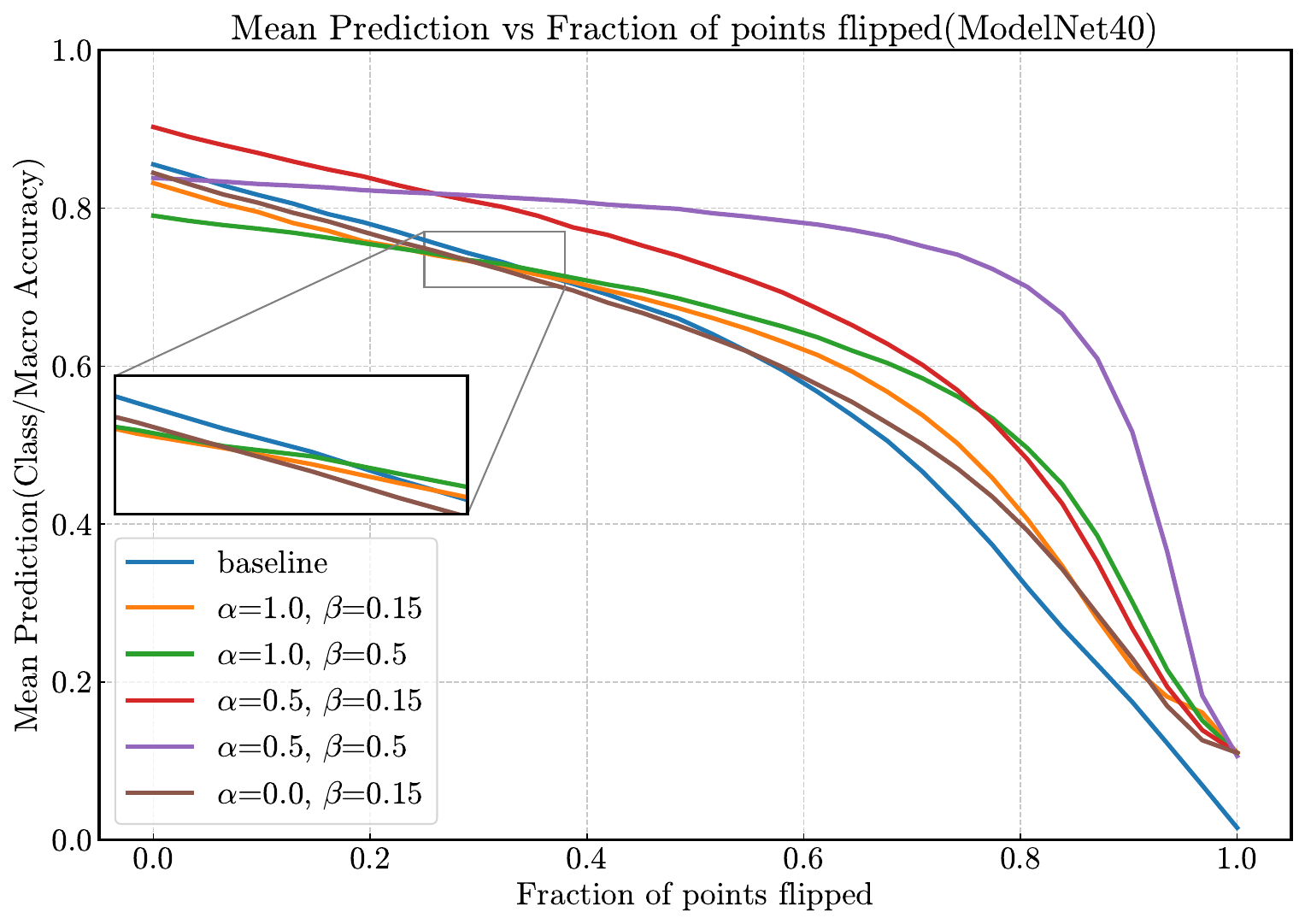}
    \caption[Point Flipping Results for PointNet++ on ModelNet-40]{Robustness of \gls{reldrop}-trained models to ordered point removal on ModelNet40 \citep{WuSKYZTX15}. \gls{reldrop}-trained models are more robust against point removal, especially when a balance between random and relevance-guided input data augmentation is employed ($\alpha=0.5$ and $\beta=0.5$).}
    \label{fig:Quantus_Results_ModelNet}
\end{figure}

The results of this experiment are shown in Table \ref{table:PointNet++_results_table}. Here, the models trained with an equal combination of random and \gls{reldrop}, i.e., $\alpha=0.5$, consistently outperform both the baseline and the models utilizing fully random input data augmentation ($\alpha=1.0$). For both the ModelNet40 and the ShapeNet datasets, we observe the parameter configurations $\alpha=0.5$ and $\beta=0.15$ to perform the best. Compared to the baseline, we observe an average overall improvement of $0.74\%$ and $0.61\%$ for the ModelNet40 and ShapeNet datasets in \emph{Class} accuracy. The fully attribution-guided models ($\alpha=0.0$) are outperformed by the combined models ($\alpha=0.5$), which aligns with our above hypothesis (cf. Section \ref{section:RelDrop}) about the potentially detrimental effects of consistently occluding the most important features. Interestingly, however, the fully attribution-guided models outperform the fully random-guided models ($\alpha=1.0$) for the best value of $\beta=0.15$ by a small margin. Again, these results highlight an increased generalization ability of \gls{reldrop}-trained models.

We also observe that for a fixed value of $\alpha$, increasing $\beta$ beyond $0.15$ (and thus reducing the threshold $RHS=(1-\beta)$) results in reduced instance accuracies. Since increasing $\beta$ leads to more points being removed, if $\beta$ becomes too large, the remaining features may not be sufficiently discriminative to accurately classify an object (cf. this is captured qualitatively in the Figure \ref{fig:QualModelNet40Dropout}). As discussed in Section \ref{section:RelDrop}, a smaller percentage of optimal information removal is to be expected for input occlusion, which removes information entirely from the prediction graph, in contrast to, e.g., dropout, where larger amounts of information removal ($\approx 50\%$) are feasible. Therefore, selecting an appropriate value for $\beta$ is essential. During our experiments, we found $\beta=0.15$ to perform best.

\subsubsection{Robustness towards Point Removal}

By removing or occluding (currently) important features, \gls{reldrop} aims to force models to base their inference on a larger number of features and thus increase their robustness and generalization ability. While we demonstrated increased test performance (implying improved generalization ability) when applying \gls{reldrop} in the previous section, this does not necessarily imply increased robustness, i.e. that the model utilizes more features for predicting. Therefore, we perform an ablation study in the following, using the feature perturbation \citep{Bach2015Pixel, SamekBMLM17, HedstromWKBMSLH23} metric. Originally conceptualized for image classification, this metric progressively occludes pixels from the most to least relevant according to an attribution map, evaluating the model's response. Here, we reformulate this metric as \emph{point flipping} to evaluate robustness towards occlusion by progressively removing points (setting their $(x,y,z)$ coordinates to $(0, 0, 0)$) and measuring the prediction accuracy after each step. This approach simulates real-world scenarios where objects might be partially occluded or incompletely scanned. In our experiments with point input clouds of size $N=1024$, we remove $32$ points per step, requiring 32 total steps until no points are left. We test various models trained with different \gls{reldrop} hyperparameters from Table \ref{table:PointNet++_results_table}.

The results of this experiment are displayed in Figure \ref{fig:Quantus_Results_ModelNet}. Initially, the best-performing model ($\alpha=0.5$, $\beta=0.15$, \emph{red}) maintains a mean prediction $\approx5\text{-}6\%$ higher than the baseline \emph{(blue)} until around $40\%$ of points are removed.  Beyond $30\text{-}35\%$ removal of points, the baseline drops steeply, while the others decline more gradually. At $55\%$ removal, the baseline reaches $\approx\! 55\%$, lower than all models. After $75\%$ removal, the best-performing model falls below ($\alpha=1.0$, $\beta=0.5$) \emph{(green)}. Notably, ($\alpha=0.5$, $\beta=0.5$) \emph{(purple)} starts lower but performs best after $25\%$ removal, outperforming all models towards the end. A higher retention of performance indicates a higher degree of robustness against point removal, and as such, we infer that input data augmentation results in more robust models, with \gls{reldrop} increasing this effect up to a point. The model with ($\alpha=0.5$, $\beta=0.5$), using an equal balance between random and relevance-guided input data augmentation, retains performance the longest, by a large margin.

\section{Limitations}
\label{section:limitations}
In the previous sections, we demonstrated the positive effects of applying \gls{reldrop}, such as improved model generalization and robustness towards occlusion. However, \gls{reldrop} is also subject to several limitations. For instance, training a model with the same number of parameters using \gls{reldrop} increases compute time by 2–2.5$\times$ compared to the baseline and \gls{re}. This occurs because each data batch requires an additional backward pass to compute attributions. Additionally, \gls{reldrop} introduces additional hyperparameters specific to the attribution method, which demand additional time and computing resources for optimal tuning. To balance random and attribution-based erasing, we rely on further hyperparameters that require tuning, but nevertheless make recommendations for optimal values in this work.%

\section{Conclusion and Outlook}
We propose \gls{reldrop}, an alternative to random input data augmentations as a general framework in Section \ref{section:RelDrop}. It can be extended and applied to different domains and tasks as a strategic regularizer by choosing different \gls{xai} methods and masking strategies. We evaluate our method's effectiveness for the 2D image and 3D point cloud classification tasks, choosing \gls{lrp} as the attribution method with ResNet and PointNet++ architectures and various benchmark datasets as covered in Section \ref{section:results}.  In comparison to random data augmentation, we observe double the improvement in average test accuracies against the baseline for both the 2D image and 3D point cloud classification tasks, consistent across different models and datasets. We observe a similar trend for zero-shot tests, where our method increases the model's generalization capabilities. \gls{reldrop} achieves this by nudging the model to focus on more diverse features in the input for the prediction, rendering it more robust against feature removal.

For future research, our approach can be extended beyond the ResNet and PointNet++ architectures to transformers, autoencoders, variational autoencoders, and generative models, provided corresponding attribution methods are available; e.g., extensions to attention modules \citep{AchtibatHDJWLS24}. Further exploration includes erasing multiple smaller patches, experimenting with different block/patch sizes and shapes, and assessing the impact on different architectures. Building on \gls{re} \citep{Zhong0KL020}, we study the effects of \gls{reldrop} through a localised block erasure technique that can serve as a foundation for developing \gls{reldrop}-based adaptations of CutMix \citep{YunHCOYC19}, MixUp \citep{ZhangCDL18}, and other augmentation techniques. Additionally, future studies can explore its effectiveness across diverse input modalities, different datasets, a broader range of tasks, and different attribution methods. Furthermore, balancing the strategy of random and attribution-guided augmentations beyond hyperparameter selection is subject to future work.

\section{Acknowledgements}

\scalerel*{\includegraphics{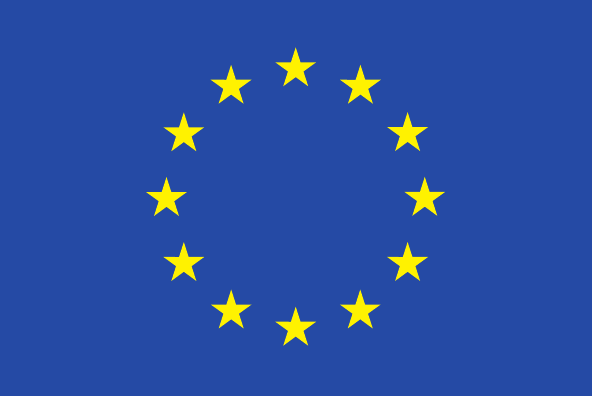}}{eu_flag.pdf} This work was supported by the European Union’s Horizon Europe research and innovation programme (EU Horizon Europe) as grants [ACHILLES (101189689), TEMA (101093003)].

This work was further supported by the Federal Ministry of Education and Research (BMBF) as grant BIFOLD (01IS18025A, 01IS180371I);  the German Research Foundation (DFG) as research unit DeSBi [KI-FOR 5363] (459422098) and the Fraunhofer Internal Programs (Fraunhofer) as grant ESPINN (PREPARE 40-08394).

{
    \small
    \bibliographystyle{bibstyle}
    \bibliography{main_20250527184904}
}
\appendix

\renewcommand\theequation{A.\arabic{equation}}
\renewcommand\thealgorithm{A.\arabic{algorithm}}
\renewcommand\thefigure{A.\arabic{figure}}
\setcounter{equation}{0}
\setcounter{figure}{0}

\section{Technical Appendix}

\subsection{Details on Attribution Computation} \label{section:BN_LRP}

\begin{figure}[h]       
    \fbox{\includegraphics[width=0.3\linewidth]{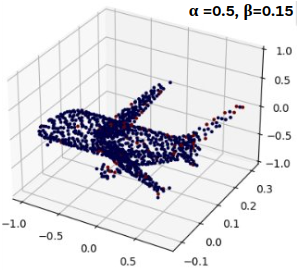}}   
    \hspace{1pt}
    \fbox{\includegraphics[width=0.3\linewidth]{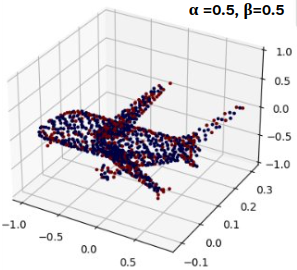}}
    \hspace{1pt}
    \fbox{\includegraphics[width=0.3\linewidth]{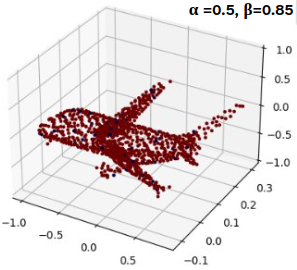}}
    \caption[ModelNet40 Qualitative attribution maps]{Attribution maps indicate the dropout strategy for different dropout parameters during training. The points highlighted in red indicate the $(x, y, z)$ coordinates of the points being dropped and replaced by $(0, 0, 0)$, and the ones highlighted in blue are unaltered and retained for the next epoch of training. An \textit{\emph{"Aeroplane"}} sample is considered for illustration with a constant value of $\alpha=0.5$ and varying values of $\beta=(0.15, 0.5, 0.85)$. \textbf{Left: }represents the dropout with parameters, $\alpha=0.5$, $\beta=0.15$, \textbf{Center: }represents the dropout with parameters, $\alpha=0.5$, $\beta=0.15$ and the, \textbf{Right: }represents the dropout with parameters, $\alpha=0.5$, $\beta=0.85$.}
    \label{fig:QualModelNet40Dropout}
\end{figure}

For \gls{lrp}, several rules exist that affect the obtained attributions \citep{Bach2015Pixel, Montavon2019Layerwise}. In our experiments, we follow recommendations of these previous works. Note that we distinguish between computing attributions used in \gls{reldrop}, which should focus on faithfulness to the model, and computing attributions for visualization, which need to be understandable:
\begin{itemize}
    \item For attribution computation during \gls{reldrop}, we apply the \emph{Flat}-rule to the first layer of each model, and either the \gls{lrp}-$\varepsilon$-rule (with $\varepsilon = 1e^{-6}$ if not stated otherwise) or the \gls{lrp}-$z^{+}$-rule to all of the remaining convolutional and linear layers of the model, as specified in Appendix \ref{appendix:hyperparameters}. For \gls{rra} and point dropout computation (Table \ref{table:ResNet_results_table_Image}, Figure \ref{fig:Quantus_Results_ModelNet}, Figure \ref{fig:QualModelNet40Dropout}), we use the $z^{+}$-version of the above rule combinations.
    \item For visualization of attributions in Figure \ref{fig:Image_Qual_results}, we use a combination of \emph{Box}-rule for the first layer, $\gamma$-rule (with $\gamma=0.25$) for the convolutional layers, and $\varepsilon$-rule for the fully connected layers. Similarly, we use a combination of $\gamma$-rule (with $\gamma=0.25$) for the convolutional layers and $\varepsilon$-rule for the fully connected layers for the intermediate attribution computation in Figure \ref{fig:relAUCResNet}.
\end{itemize}

We further canonize all batch normalization layers \citep{IoffeS15}, merging them into preceding convolutional layers, as suggested by \citep{MotzkusWL22}.
Since batch normalization consists of two consecutive linear operations, it poses an issue for modified backpropagation attribution methods such as \gls{lrp}, which are not implementation-invariant.
Canonization seeks to alleviate this issue by merging the batchnorm into a single nonlinearly activated layer, thus ensuring canonical structure of the \glspl{dnn}.

To compute attributions, we use the zennit \cite{Anders2021Software} package \href{https://github.com/chr5tphr/zennit}{\url{https://github.com/chr5tphr/zennit}}, licensed under LGPL-3. To compute \gls{rra} and point flipping metrics, we rely on the Quantus \cite{HedstromWKBMSLH23} software package \href{https://github.com/understandable-machine-intelligence-lab/Quantus/tree/main}{\url{https://github.com/understandable-machine-intelligence-lab/Quantus/tree/main}}, licensed under LGPL-3.

\subsection{Datasets}
\label{appendix:datasets}

For 2D image classification experiments, we compare our method's performance against the \emph{"\gls{re} data augmentation"} \citep{Zhong0KL020}. First, we evaluate the methods on CIFAR-10 and CIFAR-100 datasets \citep{Krizhevsky09} and extend the study to a large-scale dataset, in our case, the ImageNet-1k dataset \citep{DengDSLL009} (Custom research, non-commercial license). Further, we evaluate the zero-shot performance of the models trained on the ImageNet-R \citep{HendrycksBMKWDD21}, ImageNet-A \citep{HendrycksZBSS21}, and ImageNet-O \citep{HendrycksZBSS21} datasets. The \textbf{CIFAR-10} dataset consists of 60,000 color images of the resolution $(32\times32)$, which are distributed into 10 different classes. \textbf{CIFAR-100} dataset consists of 600 images of resolution $(32\times32)$ per class with 100 classes instead of 10. Each class contains 100 test images and 500 training images. Both of these datasets can be downloaded from their source \href{https://www.cs.toronto.edu/~kriz/cifar.html}{\url{https://www.cs.toronto.edu/~kriz/cifar.html}}. For all the pre-trained models and our experiments, we use \textbf{ImageNet-1k (ILSVRC) 2012}, the most popular ImageNet subset, which includes 1,281,167 training images covering 1,000 item classes, 100,000 test images, and 50,000 validation images. We also resize and crop each of these images to a resolution of $(224 \times 224)$ for training. The ImageNet-1k dataset can be downloaded from its source \href{https://www.image-net.org/download.php}{\url{https://www.image-net.org/download.php}}. 

\textbf{ImageNet-S} is a Large Scale Unsupervised Semantic Segmentation (LUSS) benchmark dataset that collects and annotates pixel-level labels from the ImageNet-1k dataset. Following the removal of some unsegmentable categories, such as bookshops, there are 919 categories with 1,183,322 training images and 12,419 validation images, each with its segmentation mask. It can be downloaded from \href{https://github.com/LUSSeg/ImageNet-S}{\url{https://github.com/LUSSeg/ImageNet-S}}. \textbf{ImageNet-R (rendition)} includes a test set of 30,000 images of 200 distinct classes of the original ImageNet-1k dataset, including cartoons, deviant art, graffiti, embroidery, graphics, origami, paintings, sculptures, and drawings \citep{HendrycksBMKWDD21}. The ImageNet-R dataset's real-world distribution shift, which includes variations in image style, blurriness, and other factors, may be downloaded from \href{https://people.eecs.berkeley.edu/~hendrycks/imagenet-r.tar}{\url{https://people.eecs.berkeley.edu/~hendrycks/imagenet-r.tar}} and is used to evaluate the model's generalization and out-of-distribution robustness. \textbf{ImageNet-A} examples are part of the ImageNet-1k samples; however, they are more difficult and can consistently cause classification errors across various models because of scene complexity and classifier blind spots. \textbf{ImageNet-O} (MIT license) uses out-of-distribution image examples from ImageNet-1K that models repeatedly misrepresent as high-confidence in-distribution instances. In contrast to ImageNet-A, which allows testing image classification performance when the distribution of input data changes, ImageNet-O allows testing out-of-distribution detection performance when the distribution of labels changes \citep{HendrycksZBSS21}. Both ImageNet-A and ImageNet-O can be downloaded from \href{https://people.eecs.berkeley.edu/~hendrycks/imagenet-a.tar}{\url{https://people.eecs.berkeley.edu/~hendrycks/imagenet-a.tar}} and \href{https://people.eecs.berkeley.edu/~hendrycks/imagenet-o.tar}{\url{https://people.eecs.berkeley.edu/~hendrycks/imagenet-o.tar}} respectively.

For the 3D point cloud classification experiments, we evaluate the effect of dropping out the most important points against randomly dropping out points on the two well-known datasets, ModelNet40 and ShapeNet. The \textbf{ModelNet40} dataset contains 100 unique objects per category, which are further augmented by rotating each model every 30 degrees along the gravity direction (i.e., 12 poses per model), resulting in a total of 48,000 CAD models (with rotation enlargement), of which 38,400 are used for training and 9,600 for testing \citep{WuSKYZTX15}. \textbf{ShapeNetPart (just ShapeNet in further references)} (custom non-commerical license), a portion of the entire ShapeNetCore dataset, which contains 33,700 distinct, manually aligned and annotated 3D models is used for our study \citep{ChangFGHHLSSSSX15}. Both ModelNet40 and ShapeNet can be downloaded from their respective sources, \href{https://modelnet.cs.princeton.edu/}{\url{https://modelnet.cs.princeton.edu/}} and \href{https://www.kaggle.com/datasets/mitkir/shapenet/data}{\url{https://www.kaggle.com/datasets/mitkir/shapenet/data}} respectively.

\subsection{Metrics}
\label{appendix:metrics}
\subsubsection{Accuracy}\label{subsection:Accuracy}
During our experiments, we compute \emph{micro} and \emph{macro} accuracies, where the former averages accuracy over samples, and the latter averages over classes. Micro accuracy (``instance accuracy'' or simply ``accuracy'') is measured for both the 2D and 3D settings, while we only compute macro accuracy (``class accuracy'') in the 3D setting. Note that we refer to these metrics as ``train accuracy'' if they were computed on the training data, and ``test accuracy'' if they were computed on the test or validation data --- since only validation data is available for some of the datasets, such as ImageNet-1k.

\subsubsection{\glslong{rra}} 

The \glslong{rra} quantifies the proportion of high-intensity attribution scores that fall under the ground truth segmentation mask \citep{ArrasOS22}. It can be computed by selecting the top-$K$ relevant values highlighted by the attribution method, where $K$ represents the size of the ground truth mask, \emph{\glslong{gt}}:

\begin{equation}
P_{\text{top-}K} = \{ p_1, p_2, \ldots, p_K \mid \mathcal{R}_{p_1} > \mathcal{R}_{p_2} > \ldots > \mathcal{R}_{p_K} \}
\end{equation}

Where $P_{\text{top-}K}$ is the set of $K$ pixels with $\mathcal{R}_{\text{top-}K}$  relevance scores $\mathcal{R}_{p_1}, \mathcal{R}_{p_2}, \ldots, \mathcal{R}_{p_K}$.

Then, we divide the number of these values that fall within the ground truth locations by the ground truth's size:, obtaining the \gls{rra} of a single sample as follows:

\begin{equation}
\text{\glslong{rra}} = \frac{|P_{\text{top-}K} \cap GT|}{|GT|}
\end{equation}

Taking the average \gls{rra} over all samples in the dataset yields the Mean \gls{rra} we report in our results.

\subsection{Details on Experiments}\label{appendix:exp}
\subsubsection{2D Image Classification}

\begin{algorithm}[t]
  \caption{\gls{reldrop} for 2D Images}
  \label{appendix:AlgRelDrop2D}
  \begin{algorithmic}[1]
    \Inputs{Input image $\mathcal{I}^{2d}$; Image width $W$; Image height $H$; Area of image $S$; Relevance Erasing probability $p$; Erasing area ratio range $S_{\text{low}}$ and $S_{\text{high}}$; Aspect ratio range of the Erasing block $r_{\text{low}}$ and $r_{\text{high}}$; $centroid_{\text{xy}}=(x_{\text{cen}}, y_{\text{cen}})$;}
    \Outputs{Erased Image $\mathcal{I}^{2d^{*}}$}
    \Initialize{$p_{\text{init}} \longleftarrow \text{Rand}(0,1)$}
    \If{$p_{\text{init}} \geq p$}
        \State $\mathcal{I}^{2d^{*}} \longleftarrow \mathcal{I}^{2d}$;
        \State return $\mathcal{I}^{2d^{*}}$
    \Else
        \While{True}
        \State $S_O \leftarrow Rand(S_{\text{low}}, S_{\text{high}}) * S$;
        \State $r_O \leftarrow Rand(r_{\text{low}}, r_{\text{high}})$;
        \State $H_O \leftarrow \sqrt{S_{O} * r_{O}}$, $W_O \leftarrow \sqrt{\frac{S_{O}}{r_{O}}}$;
        \State $x_{\text{cen}} \leftarrow centroid_{\text{xy}}[0]$, $y_{\text{cen}} \leftarrow centroid_{\text{xy}}[1]$;
        \Comment{$centroid_{\text{xy}}$ is the most relevant pixel}
        \If{$x_{\text{cen}} + W_{O} \leq W$ and $y_{\text{cen}} + H_{O} \leq H$}
            \State $O = \left\{ (x, y) \ | \ x \in \left[x_{\text{cen}} - \frac{W_O}{2}, x_{\text{cen}} + \frac{W_O}{2} \right], y \in \left[ y_{\text{cen}} - \frac{H_O}{2}, y_{\text{cen}} + \frac{H_O}{2} \right] \right\}$;
            \State $\mathcal{I}^{2d}(O) \leftarrow \text{Mean}(R,G,B)$;
            \State $\mathcal{I}^{2d^{*}} \longleftarrow \mathcal{I}^{2d}$;
            \State return $\mathcal{I}^{2d^{*}}$
        \EndIf
        \EndWhile
    \EndIf
  \end{algorithmic}
\end{algorithm}

\begin{algorithm}[t]
  \caption{\gls{reldrop} for 3D Point Clouds}
  \label{appendix:AlgRelDrop3D}
  \begin{algorithmic}[1]
    \Inputs{Input point cloud $\mathcal{I}^{3d} = \{(x_i, y_i, z_i)|i \in \{1, ..., N\}\}$; Point cloud size $N$; Relevance score vector $\mathcal{R}^{3d}_{norm} \in \mathbb{R}^N$; Parameters $\alpha, \beta \in [0, 1]$;}
    \Outputs{Augmented point cloud $\mathcal{I}^{3d^*}$}
    
    \State $\vec{v} \leftarrow$ Random uniform distribution $U(0,1) \in [0,1]^N$
    
    \For{each point $i \in \{1, ..., N\}$}
        \If{$\alpha\vec{v}_i + (1-\alpha)\mathcal{R}_{norm}^{3d}(i) \geq (1-\beta)$}
            \State $\mathcal{M}_{\mathcal{R}}^{3d}(i) \leftarrow 0$;  \Comment{Mark point}
            for occlusion
            \State $\mathcal{I}^{3d^*}(i) \leftarrow (0, 0, 0)$;\Comment{Replace with origin}
        \Else
            \State $\mathcal{M}_{\mathcal{R}}^{3d}(i) \leftarrow 1$ ;\Comment{Keep point}
            \State $\mathcal{I}^{3d^*}(i) \leftarrow \mathcal{I}^{3d}(i)$; \Comment{Retain the values}
        \EndIf
    \EndFor
    
    \State return $\mathcal{I}^{3d^*}$
  \end{algorithmic}
\end{algorithm}

ResNet architectures \citep{He2016Deep} of different depths (18, 34, 50) on the CIFAR-10 and CIFAR-100 \citep{Krizhevsky09} benchmark datasets and (18, 34, 50, 101) on the ImageNet-1k dataset \citep{DengDSLL009} are considered to study the effect of our method against the baseline without input augmentation, i.e., augmentation probability \textit{p=0}, and \gls{re}. Other standard regularization techniques (e.g., weight decay, label smoothing, and batch normalization) and simple data augmentation techniques (e.g., random horizontal flipping and cropping) are applied along with our method. For our experiments, all the additional scripts required to apply \gls{lrp} and place the blocks during batch training are built on top of the official implementation of \glslong{re} from \href{https://github.com/zhunzhong07/Random-Erasing/tree/master}{\url{https://github.com/zhunzhong07/Random-Erasing/tree/master}}. We use CIFAR-10/100 and ImageNet-1k pre-trained models as a starting point and their details can be found from \href{https://huggingface.co/edadaltocg}{\url{https://huggingface.co/edadaltocg}} (MIT license) and \href{https://huggingface.co/docs/hub/en/timm}{\url{https://huggingface.co/docs/hub/en/timm}} (Apache 2.0 license), and these are further finetuned for 100 and 50 epochs respectively with the different augmentation strategies. The finetuning starts with a high learning rate, which is gradually reduced using the cosine annealing scheduler.

\subsubsection{3D Point Cloud Classification:}

While the official implementation of the PointNet++ \citep{QiYSG17} paper uses Tensorflow, we instead utilize the alternate \textit{\emph{"PyTorch Implementation of PointNet++"}} \citep{PytorchPointnet++} for all our experiments due to the ease of setting up and running the experiments. The GitHub repository of the PyTorch implementation also contains a comparison table of the model performance with the official implementation. It is found to be reliable and matches the results of the official implementation. All the experiments reported in Table \ref{table:PointNet++_results_table} are conducted by training the PointNet++ model from scratch for 50 epochs with different dropout parameters.

\subsubsection{Hyperparameters}
\label{appendix:hyperparameters}
Table \ref{table:merged_hyperparameters} shows the final hyperparameters for ResNet and PointNet++ architectures trained on various benchmark datasets. The values under the ResNet further consist of CIFAR/ImageNet-1k variations; only one value means CIFAR and ImageNet-1k share the parameter.

\begin{table}[h]
\centering
\large %
\caption{Hyperparameters for ResNet and PointNet++ models.}
\begin{NiceTabular}{l|l|l}[hvlines]
  \Block{1-1}{Hyperparams} & \Block{1-1}{ResNet} & \Block{1-1}{PointNet++} \\
  \Block{5-1}{Optimizer} & \Block{1-1}{SGD} & \Block{1-1}{AdamW} \\
  & $lr = 0.01$ & $lr = 0.001$ \\
  & $beta = 0.9$ & $betas = (0.9, 0.999)$ \\
  & - & $eps = 1e \mhyphen 8$ \\
  & $w\_decay = 1e \mhyphen 4 / 5e \mhyphen 4$ & $w\_decay = 1e \mhyphen 4$ \\
  \Block{3-1}{Scheduler} & \Block{1-1}{CosLR} & \Block{1-1}{StepLR} \\
  & $lr_{min} = 1e \mhyphen 6$ & $step = 20$ \\
  & $T_{max} = 100 / 10$ & $\gamma = 0.7$ \\
  Batch Size & $128$ & $24$ \\
  \# points & - & $1024$ \\
  Max erase area & $s_{high} = 0.4$ & - \\
  Aspect erase & $r_{low} = 0.3$ & - \\
  LRP Composite & $\varepsilon$ $\mhyphen$ rule & $z^{+} \mhyphen$ rule \\
  $\varepsilon$ value & $0.001$ & - \\
\end{NiceTabular}
\label{table:merged_hyperparameters}
\end{table}

\subsubsection{Hardware}
\label{appendix:hardware}

Of all the experiments in this work, the Zero-shot testing in Section \ref{subsection:Mean_RRA}  ran on a \emph{local}
machine, while all other experiments ran on an internal \emph{HPC Cluster}. All deep learning models ran on the respective GPUs.

The local machine used Ubuntu 20.04.6 LTS, an NVIDIA TITAN RTX Graphics Card with 24GB of memory, an Intel Xeon CPU E5-2687W V4 with 3.00GHz, and 32GB of RAM. 

The HPC-Cluster used Ubuntu 18.04.6 LTS, an NVIDIA Ampere A100 Graphics Card with 40GB of memory, an Intel Xeon Gold 6150 CPU with 2.70GHz, and 512GB of RAM. Apptainer was used to containerize experiments on the cluster.

Estimated runtime for each experimental run was 2.5 GPU days, with a combined runtime of \mbox{$\approx$ 2500-3000} GPU days. Including preliminary and failed experiments, total runtime is estimated to 3400 GPU days. Note that 1 GPU day = 1*(NVIDIA Ampere A100) used for 24 hours.

\subsection{Additional Results}
\label{appendix:MoreResults}

\begin{figure}[!t]       
      \centering
	   \begin{subfigure}{\linewidth}
		\includegraphics[width=\linewidth]{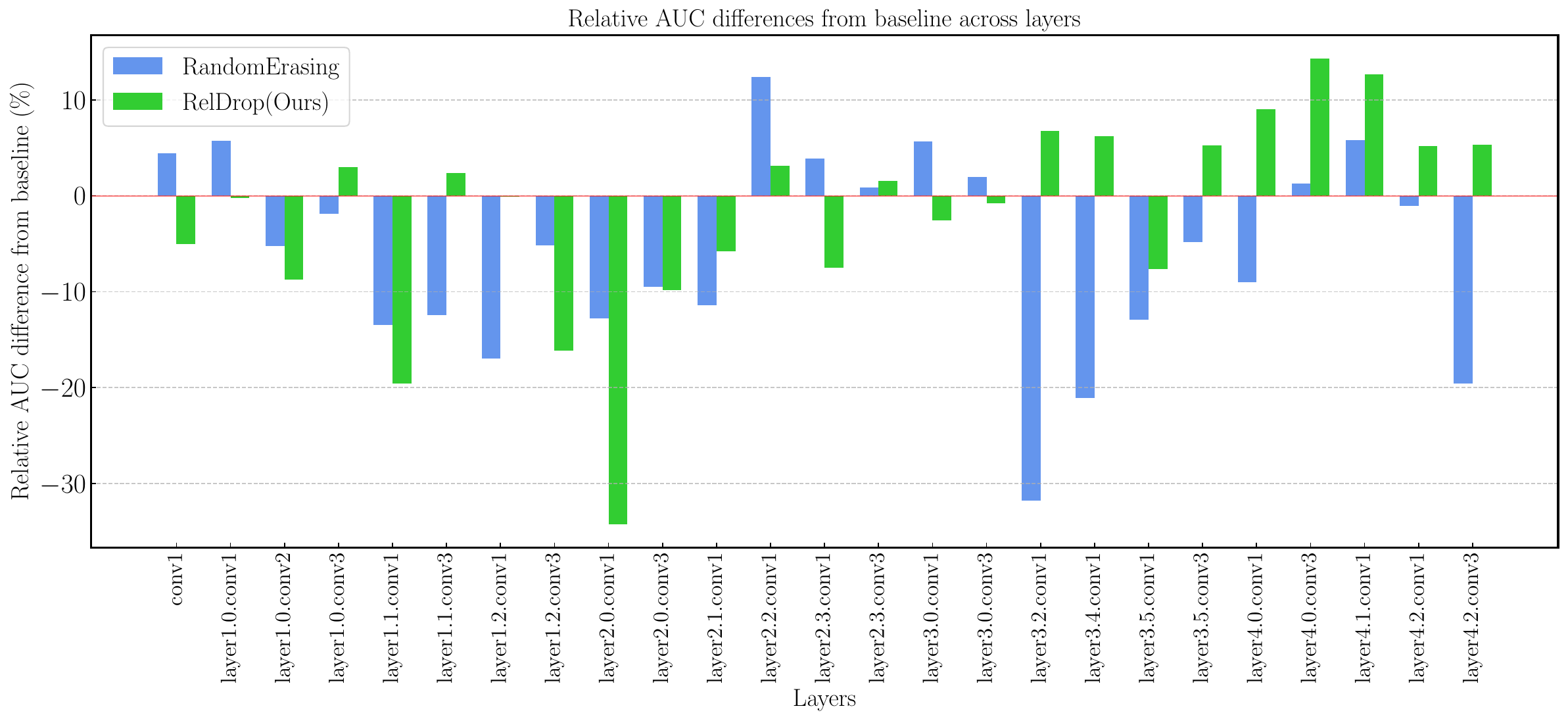}
		\label{fig:relAUCResNet50}
	   \end{subfigure}
	\vfill
	   \begin{subfigure}{\linewidth}
		\includegraphics[width=\linewidth]{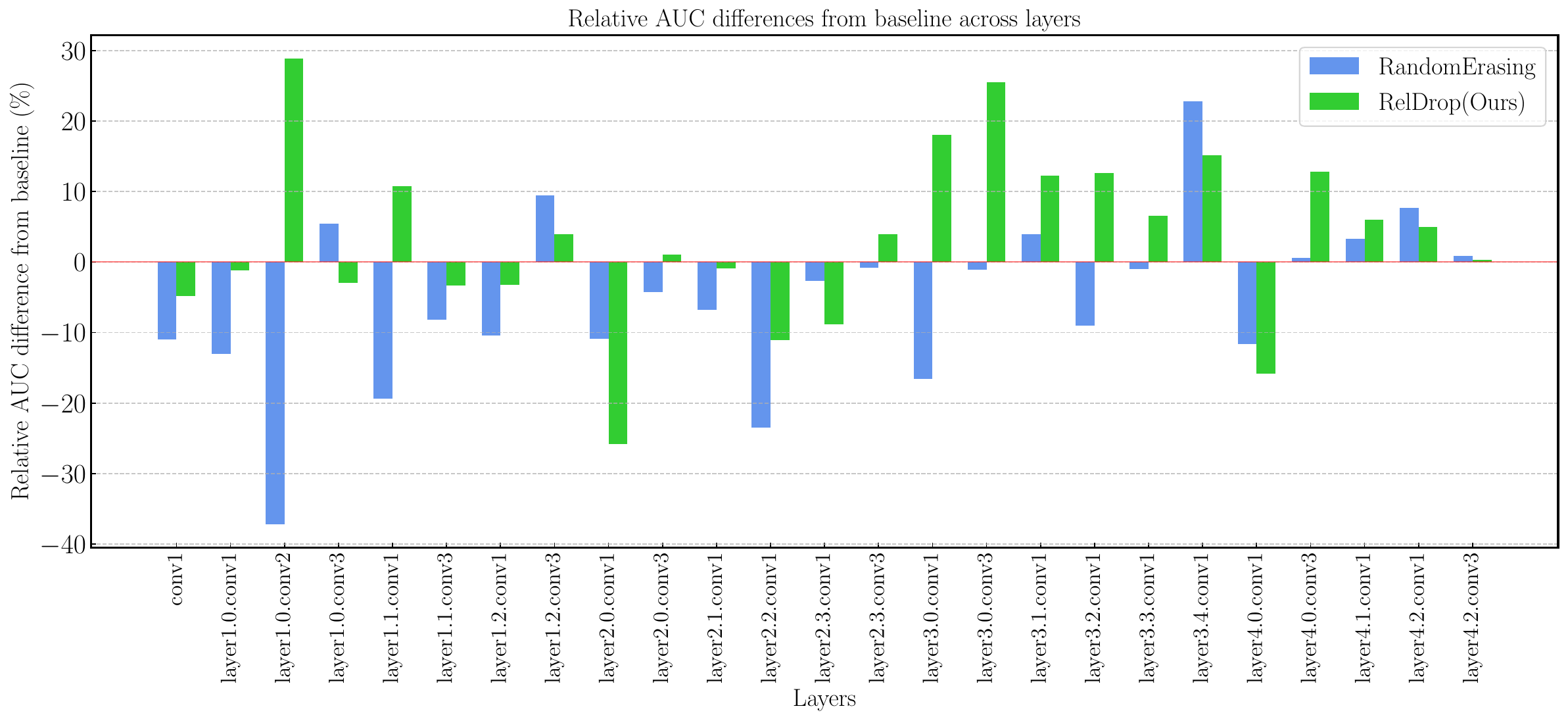}
		\label{fig:relAUCResNet101}
	   \end{subfigure}
    \caption[Relative AUC difference from baseline]{The graphs present relative AUC (\%) improvements over the baseline for \gls{re} \citep{Zhong0KL020} and \gls{reldrop}. AUC is computed by sorting channel relevance scores in descending order for each layer and normalizing by the first (maximum) value. \textbf{Top:} Relevance distribution across ResNet50 layers, ordered by network depth. \textbf{Bottom:} Relevance distribution across ResNet101 layers, also ordered by depth.}
    \label{fig:relAUCResNet}
\end{figure}

\subsubsection{Distribution of the Relevances Across Channels}
\label{appendix:RelAUCResults}
After observing the relevances being more distributed over the object of interest in Figure \ref{fig:Image_Qual_results}, we investigate this effect further, evaluating whether it is observable in intermediate layers as well. For this purpose, we compare relevance distributions at various depths of ResNet50 and ResNet101 for \gls{re} and \gls{reldrop} Figure \ref{fig:relAUCResNet}, assuming that neurons correspond to concepts \cite{Achtibat2023CRP}. Here,  we compare how the AUC of the relevance distributions differs, relative to the baseline. To compute AUC, channels are first ordered by their proportion of relevance per layer and normalized by the respective first (maximum) value. Due to this sorting, a lower AUC implies that the model's predictions are based on fewer concepts, as a higher proportion of relevance is distributed to fewer neurons. This reduced feature distribution can harm semantic expressiveness, Especially in later layers where high-level complex features are formed, a low AUC indicates lower semantic expressiveness and robustness.

In the Figure, we observe that for the initial layers of ResNet50 and ResNet101, except for a few outliers, the relative AUC differences compared to the baseline are trending towards the negative for both \gls{re} and \gls{reldrop}, with increasing magnitude until around ``Layer2''. 
This implies that in shallow layers, where the network learns extremely low-level features such as edges and textures, both methods lead to a higher concentration of relevance on fewer channels.

Throughout the deeper blocks (``Layer3'' and ``Layer4''), particularly for ResNet50, \gls{re} consistently causes large negative AUC changes over baseline, indicatingsemantically sparse representations, which may be undesirable for robustness and adaptability of the model to complex inputs. In comparison, \gls{reldrop} tends to cause strong positive increases of AUC, particularly in the latter halves of the networks. This suggests a less sparse distribution of relevance across channels and reliance on a  larger number of concepts for predicting, which is preferable for generalization and robustness. In both ResNet50 and ResNet101, \gls{reldrop} obtains significant AUC gains for the deeper layers.%
This offers an explanation as to the mechanisms causing \gls{reldrop} to consistently improve zero-shot inference performances in all the considered adversarial and distribution-shifted settings, as reported in Table \ref{table:ResNet_results_table_Image} \emph{(Green Columns)}.   

\end{document}